\documentclass[runningheads]{llncs}

\usepackage[final,year=2024]{eccv}

\usepackage{eccvabbrv}

\usepackage{graphicx}
\usepackage{booktabs}
\usepackage{marvosym}
\usepackage[accsupp]{axessibility}  

\usepackage{hyperref}
\usepackage{multirow}
\usepackage{makecell}
\usepackage{float}
\usepackage{orcidlink}

\begin{document}

\title{HPFF: Hierarchical Locally Supervised Learning with Patch Feature Fusion} 

\titlerunning{HPFF: Hierarchical Locally Supervised Learning with Patch Feature Fusion}

\author{Junhao Su \inst{1}$^*$ \and
Chenghao He \inst{2}$^*$ \and
Feiyu Zhu \inst{3}$^*$ \and
Xiaojie Xu \inst{4}$^*$ \\
Dongzhi Guan \inst{1} \and
Chenyang Si\textsuperscript{5\Letter}
}

\authorrunning{Junhao Su et al.}

\institute{Southeast University \and
East China University of Science and Technology \and
University of Shanghai for Science and Technology \and
The Hong Kong University of Science and Technology \and
Nanyang Technological University
}
\maketitle
\input{sec/0_abstract}

\section{Introduction}
\label{sec:intro}
Deep learning frameworks, such as those detailed in \cite{huang2017densely, krizhevsky2012imagenet, lecun2015deep, ren2022scaling}, predominantly employ end-to-end backpropagation \cite{rumelhart1986learning} for their learning processes. 
Despite its notable success in the realm of deep learning, it still faces several limitations. On one hand, BP presents biological implausibility \cite{crick1989recent,shen2022backpropagation} , as it relies on a global objective optimized by backpropagating error signals across layers. On the other hand, the end-to-end backpropagation process introduces the update locking problem \cite{jaderberg2017decoupled}, where hidden layer parameters cannot be updated until both forward and backward computations are completed, hindering efficient parallelization of the training process.

\begin{figure}[t]
  \centering
   \includegraphics[width=\linewidth]{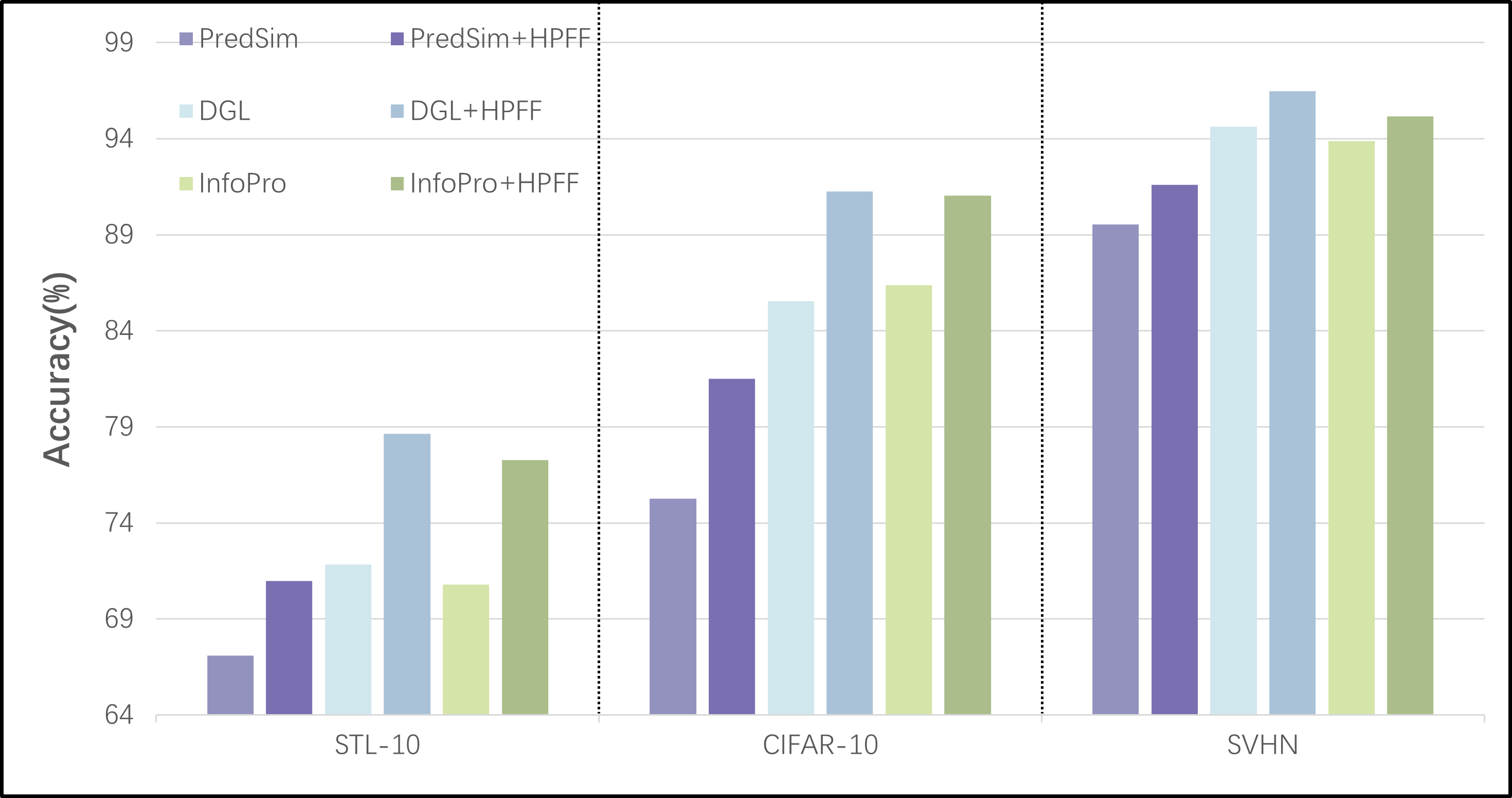}

   \caption{Comparison between different methods with HPFF and the original methods in terms of Test Accuracy. Results are obtained using ResNet-110 (K=55) on the CIFAR-10, STL-10 and SVHN datasets. The * means addtion of our HPFF.}
   \label{Figure 1}
\end{figure}

Addressing the shortcomings of backpropagation, local learning has been proposed as an alternative that divides a network into separate local modules, training each with a gradient-isolated auxiliary network to independently optimize parameters, thus avoiding the backpropagation of error signals between local modules and facilitating efficient parallelization with reduced memory usage \cite{belilovsky2019greedy,mostafa2018deep,bengio2006greedy,nokland2019training,illing2021local,wang2021revisiting}. Despite its advantages, such as alleviating the update locking problem, local learning is criticized for potentially over-focusing on local objectives, leading to suboptimal global performance due to local optima \cite{bengio2006greedy, belilovsky2019greedy}. Furthermore, due to the design of auxiliary networks, the computation process often also leads to an increase in GPU memory usage. Efforts to refine local learning, like InfoPro's refined loss function \cite{wang2021revisiting}, seek to preserve globally beneficial features, yet challenges persist with insufficient layer interaction in larger networks, often resulting in disappointing performance when scaling to networks with numerous independently optimized layers \cite{belilovsky2019greedy, nokland2019training,wang2021revisiting}.


In this paper, we introduce Hierarchical Locally Supervised Learning (HiLo) with Patch Feature Fusion (PFF). The HiLo addresses the limitations of traditional supervised local learning by employing a two-level architecture with shared weights across modules. Each level, structurally identical but operating at different granularities, generates unique losses via distinct auxiliary networks. These losses are then weighted and aggregated to update the network modules concurrently, enabling the learning of diverse characteristics that compensate for each other's deficiencies. Specifically, the cascade auxiliary network can promote information sharing to capture global features, while the independent auxiliary network is more focused on acquiring local features. We demonstrate through subsequent experiments that the independent level and cascade level complement each other. PFF divides the features within auxiliary networks into patches for computation which significantly reduces the burden on GPU memory. And by averaging and fusing features from different patches, it learns a more generalized feature representation. This method enables the network to focus on patterns that are prevalent across multiple patches.

\vspace{0.2cm}

The contributions of this paper are summarized as follows:

\begin{itemize}
    \item We propose HPFF: Hierarchical Locally Supervised Learning with Patch Feature Fusion. HPFF promoting information exchange between local modules enhances the network's ability to capture details and addresses the shortsightedness problem inherent in traditional supervised local learning methods. Simultaneously,  it significantly reduces the GPU memory occupation by auxiliary networks, further achieving savings in computational resources.
    \item HPFF serves as a versatile, plug-and-play approach within the domain of supervised local learning, applicable across various local learning methods.
    \item HPFF significantly boosts supervised local learning efficacy which can be seen in Fig. \ref{Figure 1}, achieving state-of-the-art results with notable GPU memory savings relative to end-to-end techniques.
\end{itemize}

\section{Related Work}
\label{sec:formatting}

\subsection{Local Learning}
Local learning has been proposed in recent years as an alternative training method to overcome the limitations of End-to-End (E2E) training. It primarily focuses on learning and updating parameters in gradient-isolated local modules, rather than using global error signals as in global backpropagation \cite{rumelhart1985learning}. Supervised local learning primarily enhances the learning ability of hidden layers in local modules through some handcrafted auxiliary networks \cite{belilovsky2019greedy,wang2021revisiting}, as well as some ingeniously designed local losses \cite{nokland2019training,wang2021revisiting} to retain as many global features as possible, which is beneficial for subsequent local module learning. However, the performance of these methods is relatively ordinary, with a significant gap compared to BP. In addition to supervised local learning, some works use self-supervised loss in local learning \cite{illing2021local}. For instance, Contrastive Predictive Coding (CPC) loss is used in Greedy InfoMax (GIM) \cite{lowe2019putting}, SimCLR loss is used in Local Contrastive learning (LoCo) \cite{xiong2020loco}, and Barlow Twins loss is used in Blockwise Self-Supervised Learning (Block-SSL) \cite{siddiqui2023blockwise}. These methods use self-supervised loss to ensure that a large number of features will not be easily lost when propagating between local modules. However, these methods only work when the network is divided into a small number of local modules, resulting in memory consumption that does not have a significant advantage over BP. The performance when the network is divided into a larger number of local modules remains to be verified. Another recent work \cite{journe2022hebbian} has studied local Hebbian learning, but its performance on large datasets like ImageNet \cite{deng2009imagenet} is disappointing.

\subsection{Alternatives of backpropagation}
In recent years, alternative training methods to BP have been widely researched \cite{lillicrap2020backpropagation}. Target propagation methods \cite{le1986learning,lee2015difference,bartunov2018assessing} attempt to directly propagate targets during training and use local reconstruction targets to construct special backward connections to avoid BP. On the other hand, feedback alignment \cite{lillicrap2014random,nokland2016direct} directly propagates the global error signal to each hidden layer to directly update their weights. Decoupled Neural Interfaces (DNI) \cite{jaderberg2017decoupled} avoid global gradient propagation by designing auxiliary networks to generate synthetic gradients. Some recent works \cite{dellaferrera2022error,ren2022scaling} have proposed replacing backpropagation entirely with forward gradient learning. Additionally, some methods \cite{huo2018decoupled,huo2018training} solve the update locking problem of BP and achieve parallel training by using stored previous gradients to replace current gradients for updating network parameters. However, these methods are fundamentally different from our approach of training networks with local objectives. They mostly still rely on global objectives, lack biological plausibility, and face challenges in performance on large-scale datasets \cite{deng2009imagenet}.

\begin{figure*}[t]
  \centering
   \includegraphics[width=0.95\linewidth]{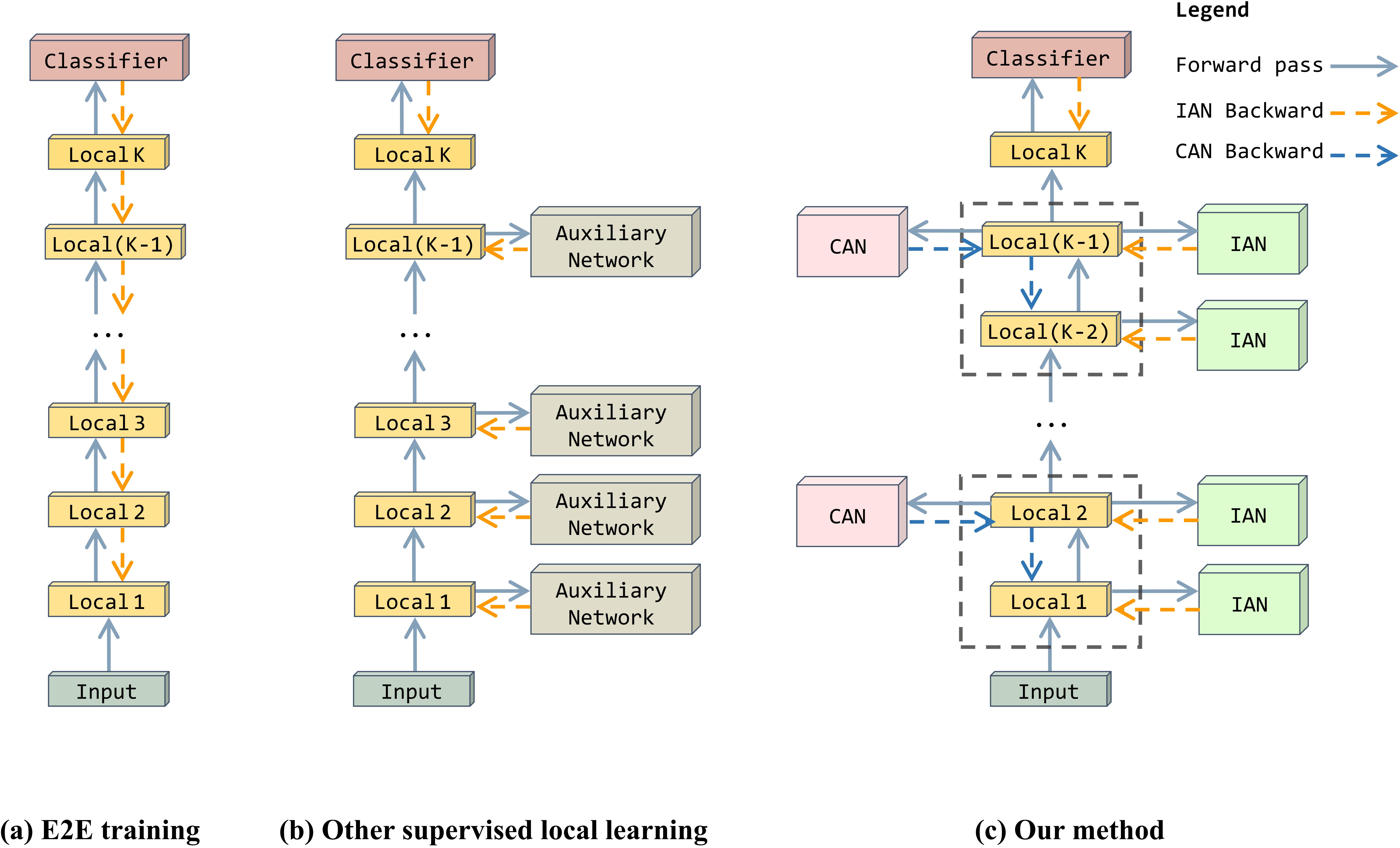}

   \caption{The HPFF overall architecture. 
   Where (a) is the structure diagram of E2E training, (b) is the structure diagram of other supervised local learning, (c) is the structure diagram of HPFF. We divide the network into K local modules in the figure. IAN stands for Independent Auxiliary Network, while CAN refers to the Cascade Auxiliary Network.}
   \label{Figure 2}
\end{figure*}

\section{Method}

\subsection{Preliminaries}
To commence, we provide an overview of the conventional end-to-end BP supervised learning paradigm and the backpropagation mechanism for contextual clarity. Suppose $x$ and $y$ denote a data sample and its corresponding ground-truth label, respectively. We represent $f_\theta$ as a network furnished with its parameters $\theta$ and its forward calculation designated as $f(\cdot)$. This deep network can be compartmentalized into multiple local modules.

During a forward pass, the output from the j-th module acts as the input for the subsequent (j+1)-th module, which we denote as $x_{j+1}=f_{\theta_j}(x_j)$. The loss function ${\mathcal{L}}(\hat{y}, y)$ is computed by contrasting the output of the final module with the ground-truth label, after which the computed loss is iteratively backpropagated to preceding modules.

Existing local learning methodologies \cite{belilovsky2019greedy,nokland2019training,wang2021revisiting} have augmented learning proficiencies through a straightforward partitioning of the network into modules, coupled with the design of an auxiliary network for each module. The output of a local module is channeled to its corresponding auxiliary network, generating the local supervision signal as $\hat{y_{j}}=g_{\gamma_j}(x_{j+1})$.

In this configuration, the parameters of the j-th auxiliary network and local module $\gamma_j, \theta_j$ are updated according to the following equations:

\begin{equation}
\gamma_j \leftarrow \gamma_j - \eta_a \times \nabla_{\gamma_j} \mathcal{L}(\hat{y_j}, y)
\end{equation}

\begin{equation}
\theta_j \leftarrow \theta_j - \eta_l \times \nabla_{\theta_j} \mathcal{L}(\hat{y_j}, y)
\end{equation}

\noindent Here, $\eta_a, \eta_l$ denote the learning rates of auxiliary networks and local modules, respectively. The employment of auxiliary networks imparts each local module with gradient-isolation, thereby enabling updates via local supervision in lieu of global backpropagation.

\subsection{Hierarchical Local Modules}

Current supervised local learning methods exhibit certain limitations. Specifically, these methods operate such that each gradient-isolated local module updates independently using an auxiliary network. This approach can lead to shortsightedness due to the lack of global error signal integration. Such shortcomings may cause individual modules to settle into local optima, which, while beneficial for the module itself, may adversely impact the overall network performance. Furthermore, the absence of inter-module coordination can lead to inefficiencies and sub-optimal results. To address these issues, we propose Hierarchical Locally Supervised Learning (HiLo), a method that facilitates communication between modules.

Within our hierarchical structure, we segregate the entire network into two levels of local modules: independent and cascade levels, as illustrated in Fig. \ref{Figure 2}. Notably, every local module serves as an independent module while simultaneously being part of a cascade module. An auxiliary network is attached to each independent or cascade module.

We posit that each cascade module encapsulates $k(k > 1)$ local modules, where local modules contained within adjacent cascade modules intersect. Specifically, the first cascade module consists of ${f_{\theta_1},\ldots,f_{\theta_k}}$, the second cascade module includes ${f_{\theta_2},\ldots,f_{\theta_{k+1}}}$, and so forth. For the j-th local module $f_{\theta_j}$, it receives supervision from one independent auxiliary network $g_{\gamma_j}$ and k cascade auxiliary networks $h_{\beta_i},\ldots,h_{\beta_{i+k-1}}$, where $i = j - k + 1$. The j-th independent auxiliary network processes the output of the j-th local module, denoted as $\hat{y_j} = g_{\gamma_j}(x_{j+1})$. The i-th cascade auxiliary network processes the output of the (i+k-1)-th local module, which is the final module within the i-th cascade module, denoted as $\hat{y_i} = h_{\beta_i}(x_{i+k})$. In this scheme, local supervision of a particular local module occurs k+1 times, once from $\mathcal{L}(\hat{y_j}, y)$ and k times from $\mathcal{L}(\hat{y_i}, y),\ldots,\mathcal{L}(\hat y_{i+k-1}, y)$.

The update rule follows:

\begin{equation}
\gamma_j \leftarrow \gamma_j - \eta_d \times \nabla_{\gamma_j} \mathcal{L}(\hat{y_j}, y)
\label{eq3}
\end{equation}

\begin{equation}
\beta_i \leftarrow \beta_i - \eta_c \times \nabla_{\beta_i} \mathcal{L}(\hat{y_i}, y)
\label{eq4}
\end{equation}

\begin{equation}
\theta_j \leftarrow \theta_j - \eta_d \times \nabla_{\theta_j} \mathcal{L}(\hat{y_j}, y) - \displaystyle \sum_{n=i}^{i+k-1} (\eta_c \times \nabla_{\theta_j} \mathcal{L}(\hat{y_n}, y))
\label{eq5}
\end{equation}

\noindent where $\eta_d, \eta_c$ are the learning rates of the independent and cascade auxiliary networks. In \eqref{eq5}, the first term of the gradient descent represents supervision from the independent auxiliary network, while the latter represents supervision from the cascade auxiliary networks.

In practice, we set $k = 2$, indicating that supervision from a cascade auxiliary network impacts two consecutive local modules. If k is set too large, the training method will be very close to BP, and this will significantly increase the GPU memory usage. Moreover, the same local module will receive local supervision more times, significantly increasing the risk of overfitting. 


\subsection{Patch Feature Fusion for Auxiliary Network}
Conventional supervised local learning habitually computes by feeding the entire feature input into an auxiliary network. Such a crude training method may not only cause the model to overlook fine-grained features that are crucial for specific tasks, leading to suboptimal final performance, but also significantly increase the GPU memory occupation by auxiliary networks due to overly large features. 
To mitigate these challenges, we introduce Patch Feature Fusion (PFF), a strategy designed to enhance computation efficiency by segmenting the input features for the auxiliary network into patches, which are subsequently fused.

To be specific, in the j-th gradient-isolated local module, Patch Feature Fusion segregates the input features $x_{j+1}$ into $n\times n$ patches for processing by the j-th auxiliary network $g_{\gamma_j}$, the feature $x_{j+1}$ can be cut as:

\begin{equation}
   x_{j+1} = \{x_{j+1}^{(k, l)} : 1 \leq k \leq n, 1 \leq l \leq n\}
\end{equation}

By partitioning $x_{j+1}$, we can calculate the local supervision signal $\hat{y_{j}}$ for the j-th auxiliary network as:
\begin{equation}
    \hat{y_{j}} = \frac{\sum \limits _{k=1} ^{n} \sum \limits _{l=1} ^{n} g_{\gamma_{j}}(x_{j+1}^{(k,l)})}{n^{2}}
\end{equation}

Upon obtaining the local supervision signal $\hat{y_{j}}$, we can then update the parameters for the auxiliary network and the main network of the j-th local module by calculating $\mathcal{L}(\hat{y_j}, y)$, as shown in \eqref{eq3},\eqref{eq4},\eqref{eq5}.

In our specific experiments, we set $n$ to 2 to ensure a balance between GPU memory reduction and performance.


\begin{table*}
  \centering
  \caption{Comparison of supervised local learning methods and BP on image classification datasets.
The averaged test errors are reported from five independent trials. The {\bfseries *} means addition of our HPFF.}
  \resizebox{1\linewidth}{!}{
  \begin{tabular}{@{}lccccc@{}}
    \hline
    \multicolumn{1}{c}{\multirow{2}{*}{Dataset}} & \multicolumn{1}{c}{\multirow{2}{*}{Method}} & \multicolumn{2}{c}{ResNet-32} & \multicolumn{2}{c}{ResNet-110} \\ 
    \cline{3-6}
    & & K=8 (Test Error) & K=16 (Test Error) & K=32 (Test Error) & K=55 (Test Error) \\ \hline
    \multirow{6}{*}{\begin{tabular}{@{}c@{}}CIFAR-10 \\ (BP(ResNet-32)=6.37,\\
    BP(ResNet-110)=5.42)\end{tabular}}  
    & PredSim & 20.62 & 22.71 & 22.08  & 24.74 \\ 
    & \textbf{PredSim*} & \textbf{13.73 ($\downarrow$6.89)} & \textbf{15.74 ($\downarrow$6.97)} & \textbf{16.51 ($\downarrow$5.57)} & \textbf{18.49 ($\downarrow$6.25)} \\ 
    & DGL & 11.63 & 14.08 & 12.51 & 14.45 \\ 
    & \textbf{DGL*} & \textbf{8.08 ($\downarrow$3.55)} & \textbf{8.94 ($\downarrow$5.14)} & \textbf{8.15 ($\downarrow$4.36)} & \textbf{8.74 ($\downarrow$5.71)} \\ 
    & InfoPro & 11.51 & 12.93 & 12.26 & 13.22 \\ 
    & \textbf{InfoPro*} & \textbf{7.79 ($\downarrow$3.72)} & \textbf{8.99 ($\downarrow$3.94)} & \textbf{8.21 ($\downarrow$4.05)} & \textbf{8.96 ($\downarrow$4.26)} \\  \hline
    \multirow{6}{*}{\begin{tabular}{@{}c@{}}STL-10 \\ (BP(ResNet-32)=19.35,\\
    BP(ResNet-110)=19.67)\end{tabular}} 
    & PredSim & 31.97 & 32.90 & 32.05 & 33.27 \\
    & \textbf{PredSim*} & \textbf{28.53 ($\downarrow$3.44)} & \textbf{29.05 ($\downarrow$3.85)} & \textbf{31.04 ($\downarrow$1.01)} & \textbf{31.75 ($\downarrow$1.52)} \\ 
    & DGL & 25.05 & 27.14 & 25.67 & 28.16 \\ 
    & \textbf{DGL*} & \textbf{20.17 ($\downarrow$4.88)} & \textbf{21.77 ($\downarrow$5.37)} & \textbf{21.29 ($\downarrow$4.38)} & \textbf{21.36 ($\downarrow$6.80)} \\ 
    & InfoPro & 27.32 & 29.28 & 28.58 & 29.20 \\ 
    & \textbf{InfoPro*} & \textbf{22.18 ($\downarrow$5.14)} & \textbf{22.37 ($\downarrow$6.91)} & \textbf{21.39 ($\downarrow$7.19)} & \textbf{22.74 ($\downarrow$6.46)} \\ \hline
    \multirow{6}{*}{\begin{tabular}{@{}c@{}}SVHN \\ (BP(ResNet-32)=2.99,\\
    BP(ResNet-110)=2.92)\end{tabular}} 
    & PredSim & 6.91 & 8.08 & 9.12 & 10.47 \\
    & \textbf{PredSim*} & \textbf{5.74 ($\downarrow$1.17)} & \textbf{6.70 ($\downarrow$1.38)} & \textbf{6.92 ($\downarrow$2.20)} & \textbf{8.41 ($\downarrow$2.06)} \\ 
    & DGL & 4.83 & 5.05 & 5.12 & 5.36 \\ 
    & \textbf{DGL*} & \textbf{3.37 ($\downarrow$1.46)} & \textbf{3.59 ($\downarrow$1.46)} & \textbf{3.21 ($\downarrow$1.91)} & \textbf{3.53 ($\downarrow$1.83)} \\ 
    & InfoPro & 5.61 & 5.97 & 5.89 & 6.11 \\ 
    & \textbf{InfoPro*} & \textbf{3.91($\downarrow$1.70)} & \textbf{4.48($\downarrow$1.49)} & \textbf{4.17 ($\downarrow$1.72)} & \textbf{4.84 ($\downarrow$1.27)} \\  \hline
  \end{tabular}}
  \label{Table 1}
  
\end{table*}

\section{Experiment}
\subsection{Experimental Setup}
We conduct experiments on four widely-used datasets: CIFAR-10 \cite{krizhevsky2009learning}, SVHN \cite{netzer2011reading}, STL-10 \cite{coates2011analysis}, and ImageNet \cite{deng2009imagenet}, utilizing ResNets \cite{he2016deep} of varying depths as our network architectures.

In our research, we combine HPFF with three advanced methods to verify its performance: PredSim \cite{nokland2019training}, DGL \cite{belilovsky2019greedy}, and InfoPro \cite{wang2021revisiting}.
Initially, each network is divided into K local modules, each consisting of an approximately equal number of layers, along with corresponding auxiliary networks.
Our proposed HPFF is applied in the training process of the first (K-1) modules, while the last module is directly connected to the global pooling layer and the fully connected layer to output the classification results.
These configurations are then benchmarked against global BP and original supervised local learning methods to ensure the removal of confounding variables for a more scientific and reasonable experiment.

\subsection{Implement Detail}
In our experiments with ResNet-32 \cite{he2016deep} and ResNet-110 \cite{he2016deep} on CIFAR-10 \cite{krizhevsky2009learning}, SVHN \cite{netzer2011reading}, and STL-10 \cite{coates2011analysis} datasets, we utilize the SGD optimizer \cite{keskar2017improving} with Nesterov momentum \cite{dozat2016incorporating} set at 0.9 and an L2 weight decay factor of 1e-4.
When we utilize ResNet-32 as the backbone network, we partition it into 8 and 16 modules. Similarly, when employing ResNet-110 as the backbone, we segment the network into 32 and 55 modules. Each local module's auxiliary network possesses its own distinct parameters.

Additionally, we use different hyperparameters on different datasets.
We employ batch sizes of 1024 for CIFAR-10 and SVHN and 128 for STL-10.
The training duration spans 400 epochs, starting with initial learning rates of 0.8 for CIFAR-10 / SVHN and 0.1 for STL-10, following a cosine annealing scheduler \cite{loshchilov2016sgdr}.
In our ImageNet \cite{deng2009imagenet} experiments, we follow different training settings for various architectures. VGG13 \cite{simonyan2014very} is trained for 90 epochs with an initial learning rate of 0.025.
For ResNet-101 \cite{he2016deep}, ResNet-152 \cite{he2016deep}, and ResNeXt-101, 32$\times$8d \cite{xie2017aggregated}, we train them for 90 epochs as well, with initial learning rates of 0.05, 0.05, and 0.025, respectively.
Batch sizes are set to 64 for VGG13 \cite{simonyan2014very}, 128 for ResNet-101 and ResNet-152, and 64 for ResNeXt-101, 32$\times$8d.

\subsection{Comparison with the SOTA results}

\noindent {\bfseries Results on various image classification benchmarks:} We first conduct experiments on CIFAR-10 \cite{krizhevsky2009learning}, STL-10 \cite{coates2011analysis}, and SVHN \cite{netzer2011reading} to verify the performance of HPFF. The experimental results are shown in Table \ref{Table 1}. It can be seen that the enhancement effect of our proposed HPFF is very obvious in all kinds of datasets and different methods, exceeding the original methods.

Experimental results on the CIFAR-10 dataset show that the more local modules the network is divided into, the more obvious the improvement brought by our method. For example, in the case of ResNet-32 (K=16), where each hidden layer is considered as a local module, HPFF can reduce the test error of PredSim, DGL, and InfoPro from 22.71, 14.08, 12.93 to 15.74, 8.94, 8.99. When using a deeper network, ResNet-110 (K=55), HPFF still brings a performance improvement of 25\%, 39\%, and 32\% for PredSim, DGL, and InfoPro respectively. Through these experimental results, we can see the universality of our HPFF. Its improvement on network performance is not limited by the depth of the network and the method.

When the experiments are extended to other datasets, HPFF still demonstrates strong performance. On the STL-10 dataset, the performance improvements brought by HPFF for PredSim, DGL, and InfoPro are more than 5\%, 17\%, and 18\%. On the SVHN dataset, the corresponding performance improvements are all over 17\%, 28\%, and 20\%. These experiments demonstrate the astonishing performance of HPFF. Even when every hidden layer of the network is treated as a gradient-isolated local module, HPFF can greatly bridge the performance gap that exists between current supervised local learning methods and BP.
\begin{table}[H]
  \centering 
  \caption{Results on the validation set of ImageNet.}
   \scalebox{0.925}{\begin{tabular}{@{}cccc@{}}
    \hline
    Backbone & Method & Top1-Error & Top5-Error \\ \hline
    \multirow{3}{*}{\begin{tabular}{@{}c@{}} VGG13 \\ (K=10) \end{tabular}}  
    & BP & 28.41 & 9.63 \\ 
    & DGL & 35.60 & 14.20 \\ 
    & \textbf{DGL*} & \textbf{32.49 ($\downarrow$ 3.11)} & \textbf{11.22 ($\downarrow$ 2.98)} \\  \hline
    \multirow{3}{*}{\begin{tabular}{@{}c@{}}ResNet-101 \\ (K=4) \end{tabular}}
    & BP & 22.03 & 5.93 \\ 
    & InfoPro & 22.81 & 6.54 \\
    & \textbf{InfoPro*} & \textbf{21.14 ($\downarrow$ 1.67)} & \textbf{5.49 ($\downarrow$ 1.05)} \\ \hline
    \multirow{3}{*}{\begin{tabular}{@{}c@{}}ResNet-152 \\ (K=4)\end{tabular}}
    & BP & 21.60 & 5.92 \\ 
    & InfoPro & 22.93 & 6.71 \\ 
    & \textbf{InfoPro*} & \textbf{20.99 ($\downarrow$ 1.94)} & \textbf{5.29 ($\downarrow$ 1.42)} \\ \hline 
    \multirow{3}{*}{\begin{tabular}{@{}c@{}}ResNeXt-101, 32$\times$8d \\ (K=4)\end{tabular}}
    & BP & 20.64 & 5.40 \\ 
    & InfoPro & 21.69 & 6.11 \\ 
    & \textbf{InfoPro*} & \textbf{19.94 ($\downarrow$ 1.75)}  & \textbf{5.09 ($\downarrow$ 1.02)} \\ \hline 
  \end{tabular}}
  \label{Table 2}
\end{table}
\noindent {\bfseries Results on ImageNet:} We evaluate the performance of HPFF on the more challenging dataset, ImageNet \cite{deng2009imagenet}. When we use VGG13 \cite{simonyan2014very} as the backbone for the experiment, and treat every hidden layer as a local module for training, as can be seen from Table \ref{Table 2}, there is a huge performance gap between DGL \cite{belilovsky2019greedy} and BP. Such disappointing performance makes it impossible to achieve effective training on large-scale datasets. After adding HPFF, Top1-Error and Top5-Error decrease by 3.11 and 2.98 points, making the performance comparable to BP.

We further conduct experiments by using ResNet-101 \cite{he2016deep}, ResNet-152 \cite{he2016deep}, and ResNeXt-101, 32$\times$8d \cite{xie2017aggregated} as backbones, dividing them into 4 local modules. As shown in Table \ref{Table 2}, the training effect of InfoPro is found to be suboptimal at this juncture, significantly lower than the performance of BP. However, when we add HPFF, the Top-1 Error of InfoPro on ResNet-101, ResNet-152, and ResNeXt, 32$\times$8d decrease by 7.3\%, 8.5\%, and 7.9\%. The decrease in Top-5 Error is even more significant, at 16.0\%, 21.1\%, and 16.7\%, respectively. Such a significant performance improvement allows the network to surpass BP in performance even when it is divided into 4 local modules.


\noindent {\bfseries Memory Consumption:} Owing to the characteristic that gradients do not propagate between local modules, local learning can effectuate significant savings in GPU memory during the training phase. We have investigated the GPU memory usage of the three methodologies with the addition of HPFF on the CIFAR-10 \cite{krizhevsky2009learning} and ImageNet \cite{deng2009imagenet} datasets. Results are depicted in Table \ref{Table 3}.

On the CIFAR-10 dataset, when PredSim \cite{nokland2019training}, DGL \cite{belilovsky2019greedy}, and InfoPro \cite{wang2021revisiting} added HPFF, the GPU memory usage of ResNet-32 (K=16) compared to BP decrease by 33.2\%, 31.5\%, and 43.3\%. When extended to the deeper network ResNet-110 (K=55), the reduction in GPU memory becomes more pronounced, reaching 73.7\%, 74.3\%, and 79.5\%. In the GPU memory test on the ImageNet, with the incorporation of HPFF into InfoPro, the GPU memory usage of ResNet-101, ResNet-152, and ResNeXt, 32$\times$8d compared to BP decrease by 19.5\%, 20.9\%, and 19.8\% respectively. This means that we have achieved higher performance with less GPU memory than BP.

\subsection{Ablation Study}
\begin{table}[t]
  \centering
  \caption{Comparison of GPU memory usage between BP and other methods with HPFF on the CIFAR-10 and ImageNet datasets.}
   \scalebox{0.925}{\begin{tabular}{@{}cccc@{}}
  \hline
  Dataset & Network & Method & GPU Memory(GB) \\ \hline
\multirow{8}{*}{CIFAR-10} 
& \multirow{4}{*}{ResNet-32(K=16)} & BP & 3.37 \\
& & \textbf{DGL*} & \textbf{2.25($\downarrow$ 33.2\%)} \\
& & \textbf{InfoPro*} & \textbf{2.31($\downarrow$ 31.5\%)} \\
& & \textbf{PredSim*} & \textbf{1.91($\downarrow$ 43.3\%)} \\ \cline{2-4}
& \multirow{4}{*}{ResNet-110(K=55)} & BP & 9.26 \\
& & \textbf{DGL*} & \textbf{2.44($\downarrow$ 73.7\%)} \\
& & \textbf{InfoPro*} & \textbf{2.38($\downarrow$ 74.3\%)} \\
& & \textbf{PredSim*} & \textbf{1.90($\downarrow$ 79.5\%)} \\ \hline
\multirow{6}{*}{ImageNet} 
& \multirow{2}{*}{ResNet-101} & BP & 19.71 \\
& & \textbf{InfoPro*(K=4)} & \textbf{15.87($\downarrow$19.5\%)} \\ \cline{2-4}
& \multirow{2}{*}{ResNet-152} & BP & 26.29 \\
& & \textbf{InfoPro*(K=4)} & \textbf{20.79($\downarrow$20.9\%)} \\ \cline{2-4}
& \multirow{2}{*}{ResNeXt-101, 32$\times$8d} & BP & 19.22 \\
& & \textbf{InfoPro*(K=4)} & \textbf{15.41($\downarrow$19.8\%)} \\ \hline
\end{tabular}}
  \label{Table 3}
  
\end{table}

\begin{table}[htbp]
  \centering
  \caption{Ablation study of our method. (a) Using DGL as baseline and ResNet-32 (K=16) as backbone on the CIFAR-10 dataset. (b) Using InfoPro as baseline and ResNet-101 (K=4) as backbone on the ImageNet dataset, IL stands for Independent Level, and CL stands for Cascade Level.}
  \label{Table 4}
  \begin{minipage}{0.35\linewidth}
    \centering
    \footnotesize 
    \begin{tabular}{cccc} 
      \hline IL & CL & PFF & Test Error \\
      \hline $\checkmark$ & $\times$ & $\times$ & 14.08 \\
      $\times$ & $\checkmark$ & $\times$ & $\boldsymbol{10.51(\downarrow 3.57)}$ \\
      $\checkmark$ & $\checkmark$ & $\times$ & $\boldsymbol{9.44(\downarrow 4.64)}$ \\
      $\checkmark$ & $\checkmark$ & $\checkmark$ & $\boldsymbol{8.94(\downarrow 5.14)}$ \\
      \hline
    \end{tabular}
    \subcaption{}
  \end{minipage}
  \hfill
  \begin{minipage}{0.55\linewidth}
    \centering
    \footnotesize 
    \begin{tabular}{ccccc} 
      \hline IL  & CL  & PFF & Top1-Error & Top5-Error \\
      \hline $\checkmark$ & $\times$ & $\times$ & 22.81 & 6.54 \\
      $\times$ & $\checkmark$ & $\times$ & $\boldsymbol{22.12(\downarrow 0.69)}$ & $\boldsymbol{5.99(\downarrow 0.55)}$ \\
      $\checkmark$ & $\checkmark$ & $\times$ & $\boldsymbol{21.35(\downarrow 1.46)}$ & $\boldsymbol{5.68(\downarrow 0.86)}$ \\
      $\checkmark$ & $\checkmark$ & $\checkmark$ & $\boldsymbol{21.14(\downarrow 1.67)}$ & $\boldsymbol{5.49(\downarrow 1.05)}$ \\
      \hline
    \end{tabular}
    \subcaption{}
  \end{minipage}
\end{table}

\begin{figure}
  \centering
  \begin{subfigure}{0.27\linewidth}
    \centering
    \includegraphics[width=\linewidth,keepaspectratio, height=1\textwidth]{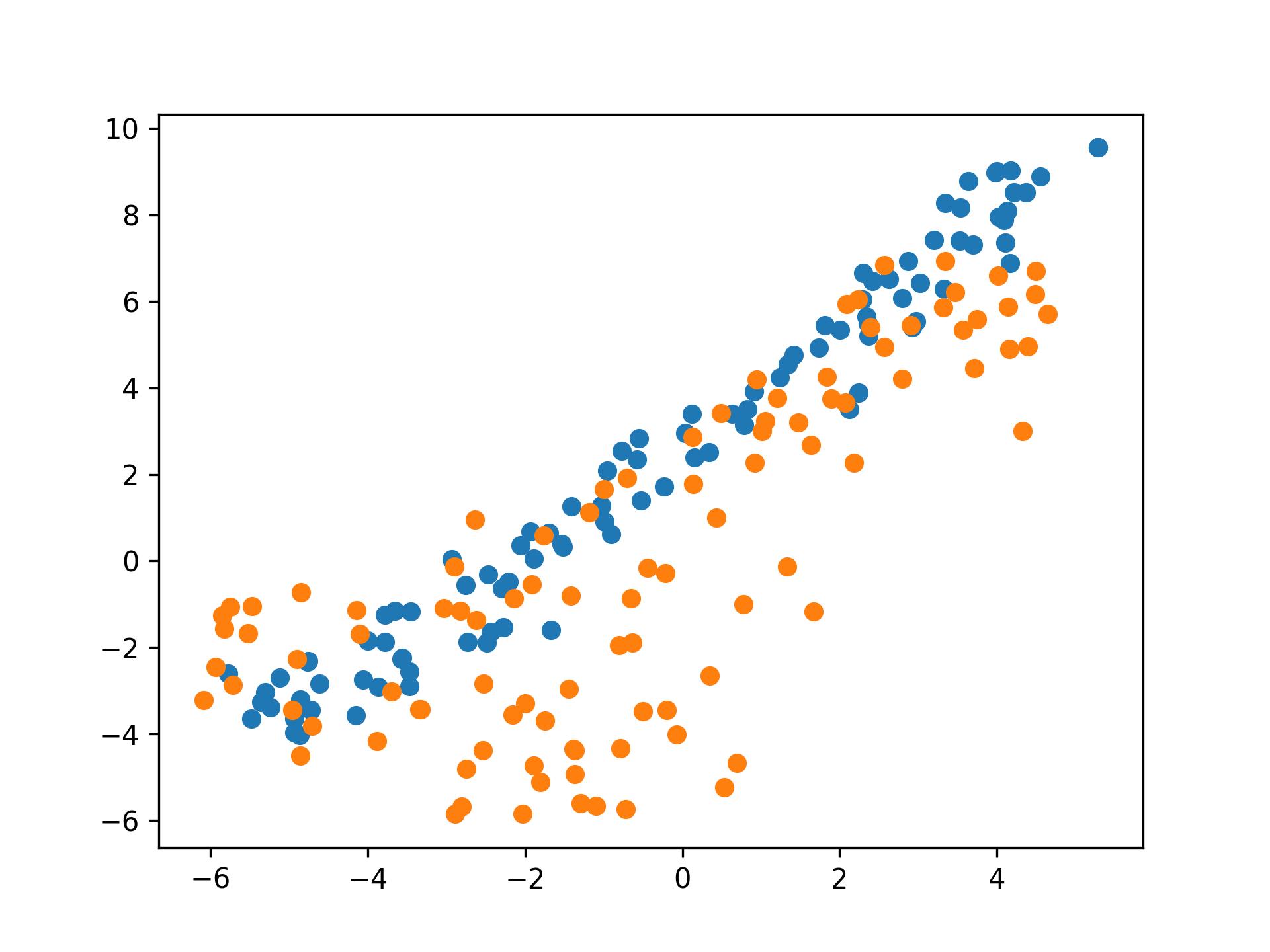}
    \caption{}
    \label{fig:short-a1}
  \end{subfigure}
  \hspace{0.02\linewidth}
  \begin{subfigure}{0.27\linewidth}
    \centering
    \includegraphics[width=\linewidth,keepaspectratio, height=1\textwidth]{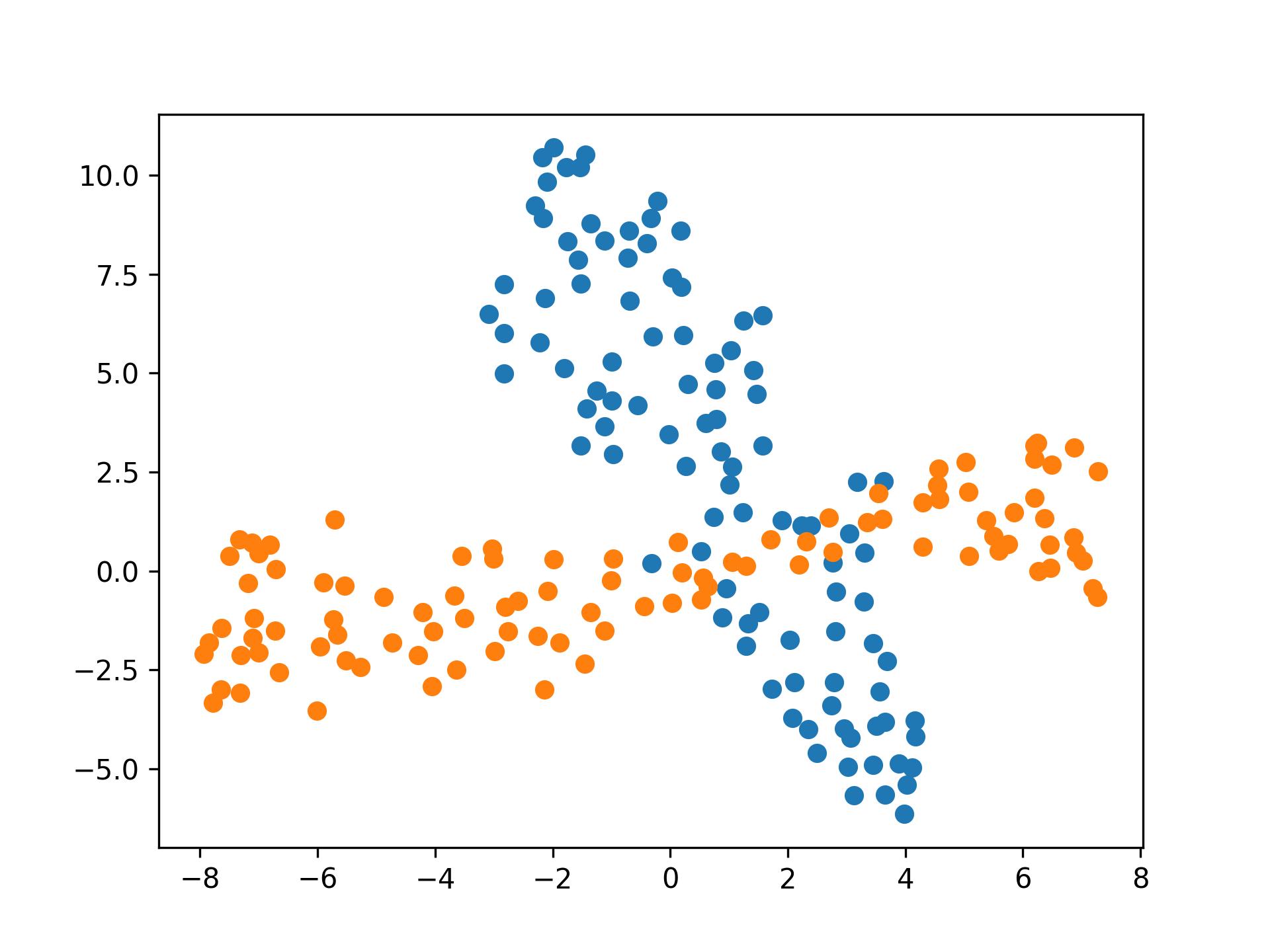}
    \caption{}
    \label{fig:short-b1}
  \end{subfigure}
  \hspace{0.02\linewidth}
  \begin{subfigure}{0.27\linewidth}
    \centering
    \includegraphics[width=\linewidth,keepaspectratio, height=1\textwidth]{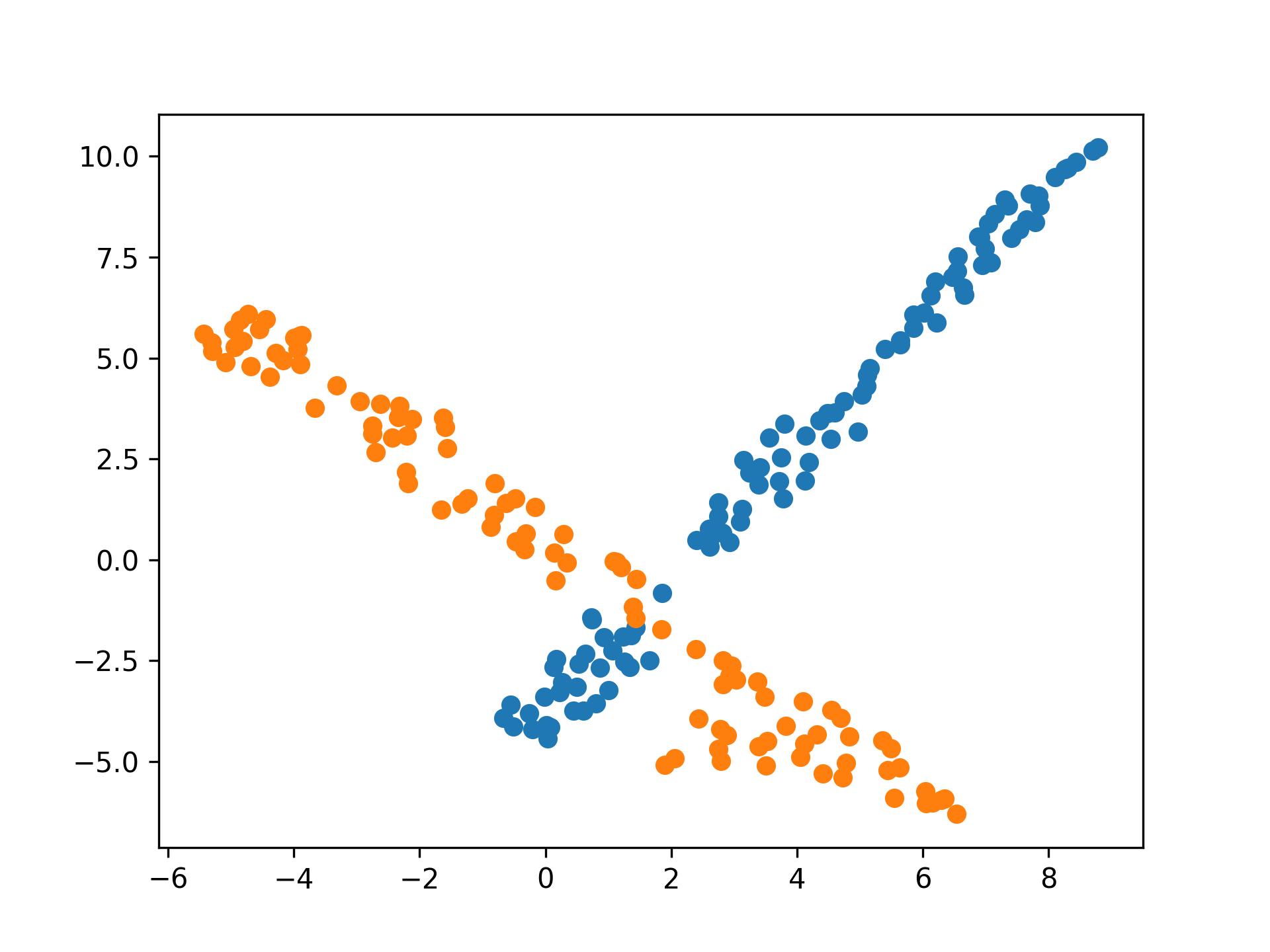}
    \caption{}
    \label{fig:short-c1}
  \end{subfigure}
  \caption{(a) is t-SNE of Independent Level, (b) is t-SNE of Cascade Level, (c) t-SNE of HiLo. Visualizations are conducted using t-SNE on the CIFAR-10 dataset, with ResNet-32 (K=16) as the backbone. The target class is represented in blue, while the non-target class is represented in yellow.}
  \label{Figure 3}
\end{figure}

\noindent {\bfseries Comparison between Independent Level, Cascade Level, and HiLo:} To further elucidate the differences between Independent and Cascade Levels in the context of feature learning, we employ t-SNE visualizations \cite{van2008visualizing}. The resulting visual representation of target and non-target points is illustrated in Fig. \ref{Figure 3}.

In the t-SNE visualization of the Independent Level, we observe that target and non-target classes are mixed together. This situation indicates that the independent level faces challenges in distinguishing global categories, resulting in a high test error of 14.08. However, data points belonging to the same class exhibit a tendency to cluster, implying that the Independent Level is proficient in learning and representing local features. These local features often appear similar or identical for objects within the same class.

Conversely, the t-SNE visualization of the Cascade Level shows a clear separation between target and non-target classes, indicating the cascade level's strong global feature learning capability and the ability to distinguish between different categories, thereby reducing the test error to 10.51. Nonetheless, the intra-class distances between data points are larger, suggesting that the Cascade Level might not optimally capture minor distinctions within the same class and may lack in local feature learning.

Fig. \ref{Figure 3}(c) presents the t-SNE visualization for HiLo method. It is evident that HiLo effectively differentiates between target and non-target classes. Furthermore, the distances between data points within the same class are significantly smaller. This indicates that HiLo successfully integrates the local feature learning capability of the Independent Level with the global feature learning ability of the Cascade Level. Consequently, HiLo not only accurately distinguishes between different categories but also captures subtle intra-class differences more effectively. This robust capability further reduces the test error to 9.44.

By analyzing the t-SNE visualizations of these three methods, we demonstrate that local and global feature learning can indeed complement each other. And HiLo embodies this principle effectively, as is intuitively demonstrated by the performance comparison in Table \ref{Table 4}.

  

\begin{figure*}[t]
  \centering
    \begin{subfigure}{0.45\linewidth}
    \centering
    \includegraphics[width=1\linewidth,keepaspectratio, height=1\textwidth]{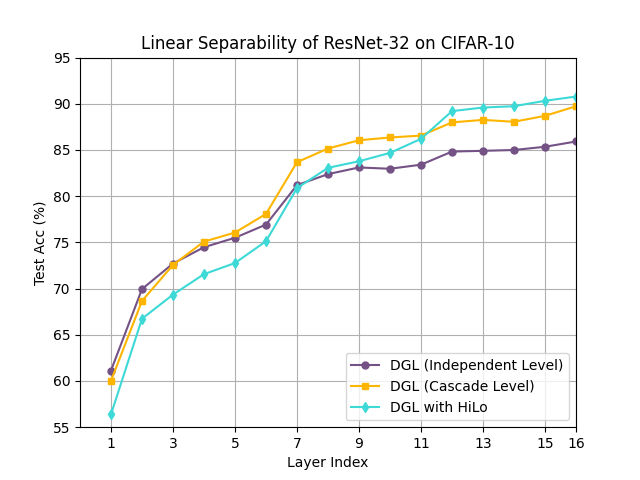}
    \label{fig:short-b}
  \end{subfigure}
  \begin{subfigure}{0.45\linewidth}
    \centering
    \includegraphics[width=1\linewidth,keepaspectratio, height=1\textwidth]{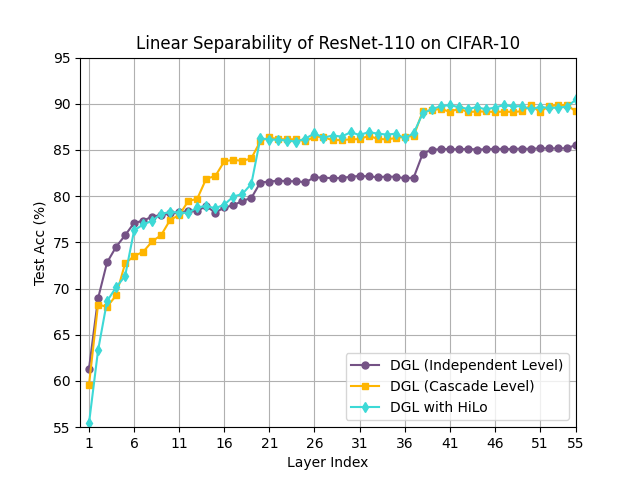}
    \label{fig:short-a}
  \end{subfigure}
   \caption{Comparsion of layer-wise linear separability across different learning rules on ResNet-32 and ResNet-110.}
   \label{Figure 4}
\end{figure*}

\begin{figure*}[t]
  \centering
  \begin{subfigure}{0.45\linewidth}
    \centering
    \includegraphics[width=1\linewidth,keepaspectratio, height=1\textwidth]{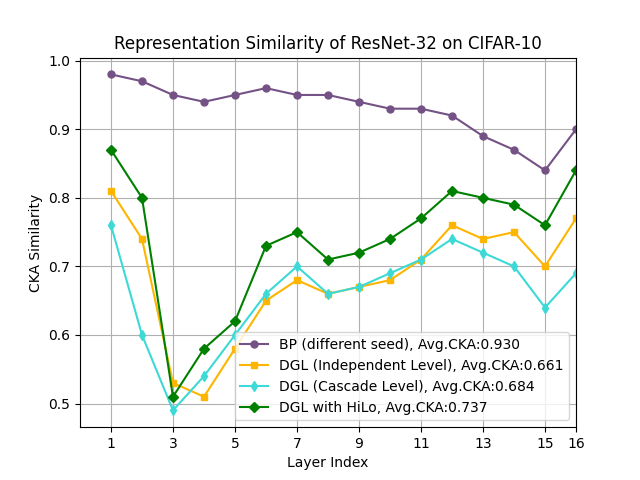}
    \label{fig:short-a3}
  \end{subfigure}
   \begin{subfigure}{0.45\linewidth}
    \centering
    \includegraphics[width=1\linewidth,keepaspectratio, height=1\textwidth]{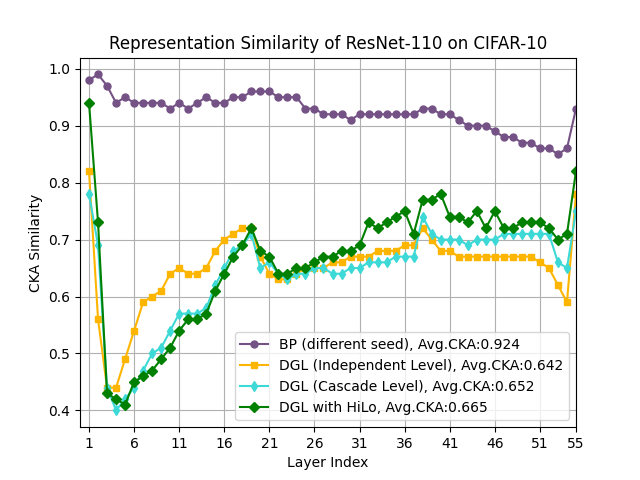}
    \label{fig:short-b3}
  \end{subfigure}
   \caption{Comparsion of layer-wise representation similarity. We utilize CKA {\cite{kornblith2019similarity}} to mesure the layer-wise similarity of representation between BP and our methods.}
   \label{Figure 5}
\end{figure*}

\noindent {\bfseries Decoupled Layer Accuracy Analysis of HiLo:} To further dissect how HiLo bolsters network performance, we freeze the parameters of the main network train a linear classifier for each gradient-isolated hidden layer to determine the classification accuracy at each layer. Fig. \ref{Figure 4} depicts this situation, where we use DGL \cite{belilovsky2019greedy} as a baseline for comparison. Upon integrating our method, we notice a significant decrease in the accuracy of the early layers, until reaching the latter layers, where the accuracy experiences a substantial increase, far exceeding the original method. We posit that high classification accuracy in early layers may not be beneficial from a global perspective. If early layers are overly focused on optimizing local objectives, they may discard certain useful features for subsequent layers. This leads to less effective information being learned by the later layers, thus resulting in suboptimal performance of the entire network. After incorporating HiLo, the classification accuracy of the early layers significantly decreases because it retains more features beneficial to the global objective, thereby enhancing the generalization capability of subsequent layers.

\noindent {\bfseries Representation Similarity Analysis of HiLo:} We conduct a CKA \cite{kornblith2019similarity} experiment to demonstrate the effectiveness of HiLo. Using DGL \cite{belilovsky2019greedy} as the baseline, we added HiLo and calculated the CKA similarity of each layer with BP \cite{rumelhart1985learning} across different methods, then averaged the results. As shown in Fig. \ref{Figure 5}, our method significantly enhances the CKA similarity of DGL, especially in the early layers. High CKA similarity in the early layers indicates a learning mode similar to BP, which helps in learning features beneficial to the global objective. The poor performance of the original method is due to the early layers focusing too much on local optimization objectives, leading to significant differences from BP and resulting in a decline in overall performance. The CKA experiment validates that HiLo successfully addresses the issues of limited information interaction and shortsightedness inherent in current supervised local learning methods.

\begin{table}[htbp]
  \centering
  \caption{Comparison between different methods in terms of test error and GPU memory usage.}
  \scalebox{0.95}{\begin{tabular}{@{}ccccc@{}}
  \hline
  Dataset & Network & Method & GPU Memory(GB)  & Test Error \\ \hline
\multirow{6}{*}{CIFAR-10} 
& \multirow{3}{*}{ResNet-32(K=16)} & InfoPro & 2.67 & 12.93 \\
& & \textbf{InfoPro+HiLo} &\textbf{3.13} & \textbf{9.59} \\
& & \textbf{InfoPro+HPFF} & \textbf{2.31} & \textbf{8.99} \\ \cline{2-5}
& \multirow{3}{*}{ResNet-110(K=55)} & InfoPro & 2.62 & 13.62 \\
& & \textbf{InfoPro+HiLo} &\textbf{3.28} & \textbf{9.35} \\
& & \textbf{InfoPro+HPFF} & \textbf{2.38} & \textbf{8.96} \\ \hline
\end{tabular}}
  \label{Table 5}
  
\end{table}

\begin{figure}[t]
\centering
	\begin{minipage}{0.22\linewidth}
		\centering
		\includegraphics[width=\linewidth]{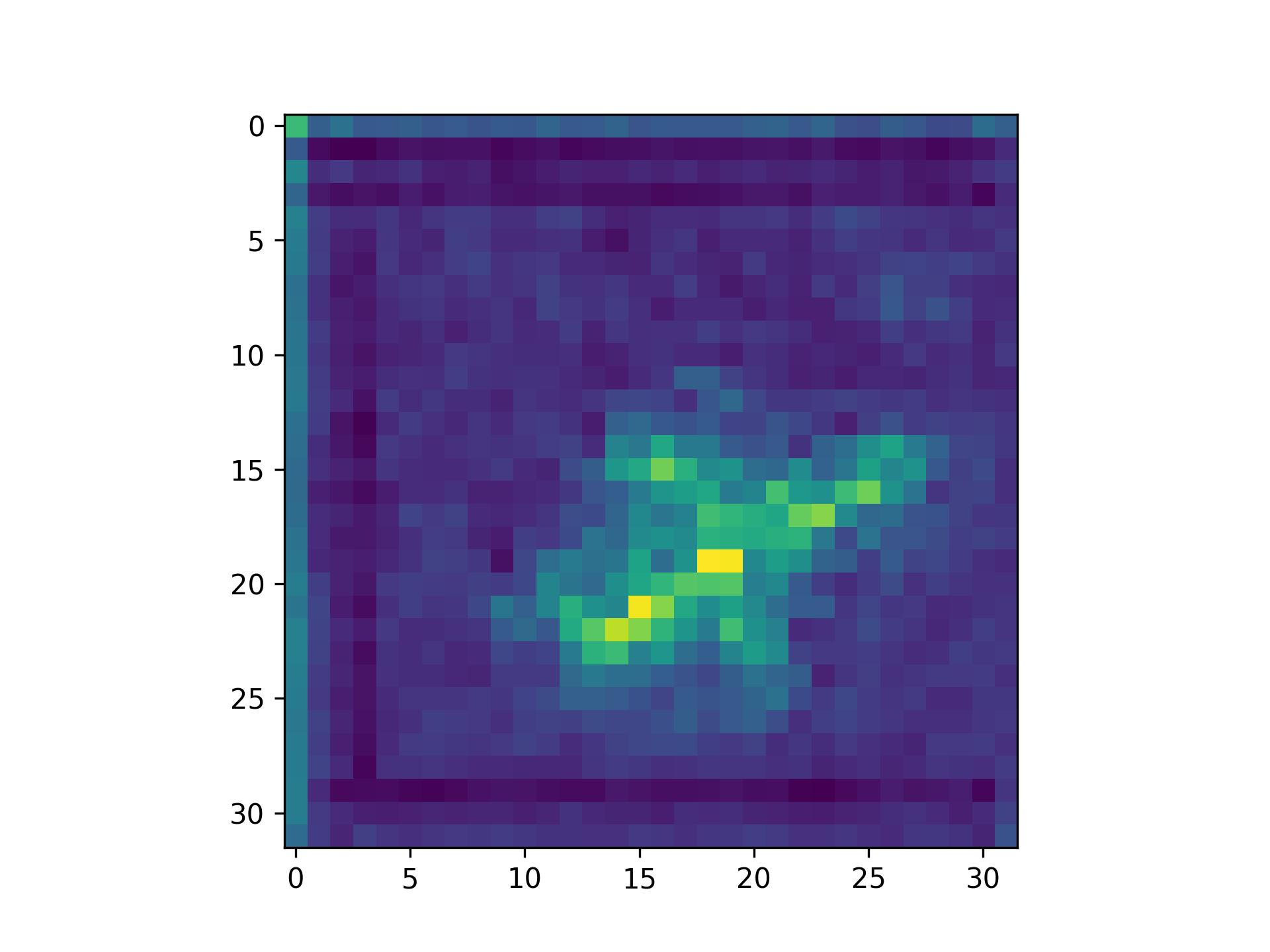}\\
  (a)
	\end{minipage}
	\begin{minipage}{0.22\linewidth}
		\centering
		\includegraphics[width=\linewidth]{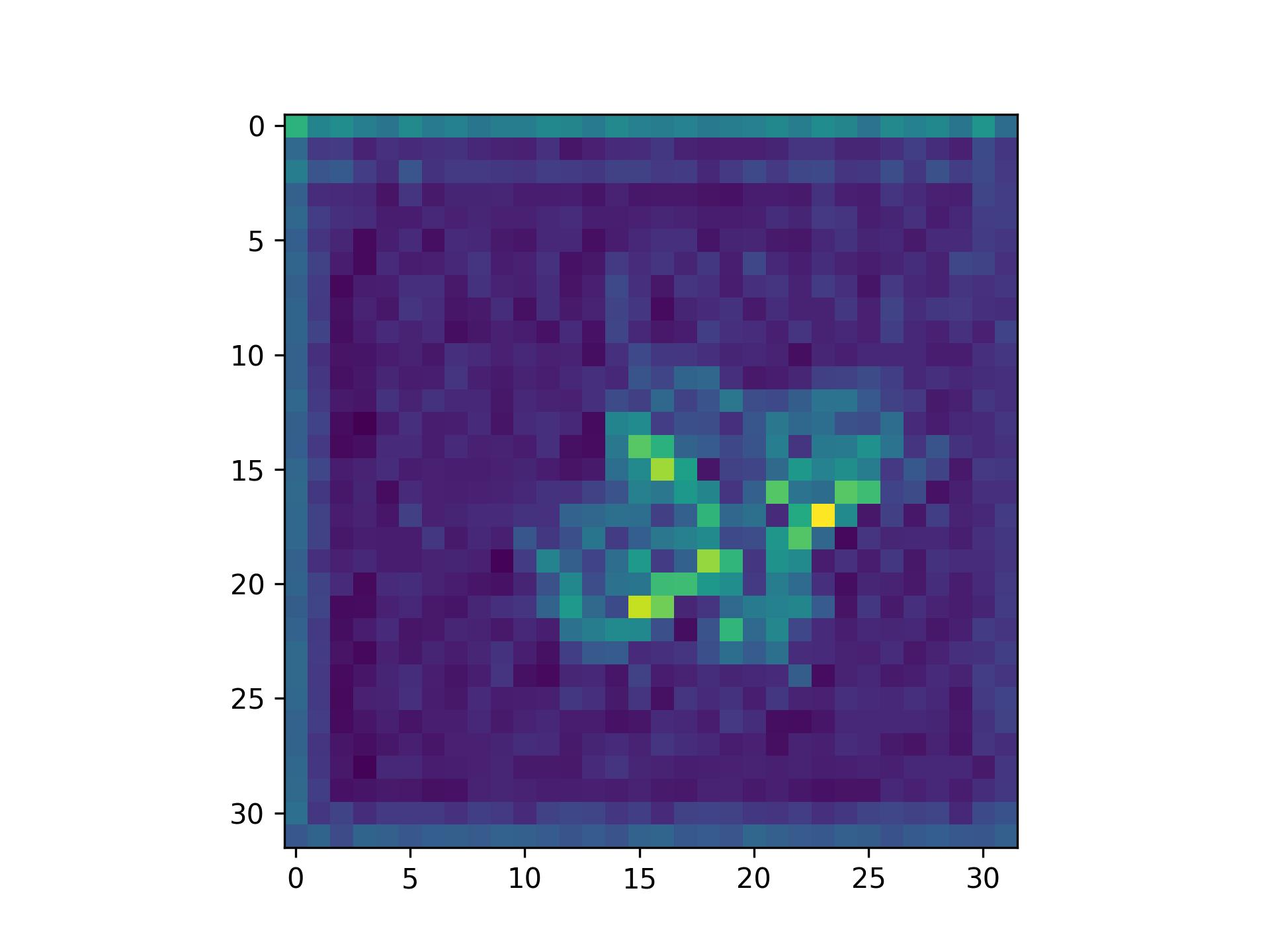}\\
  (b)
	\end{minipage}
 \begin{minipage}{0.22\linewidth}
		\centering
		\includegraphics[width=\linewidth]{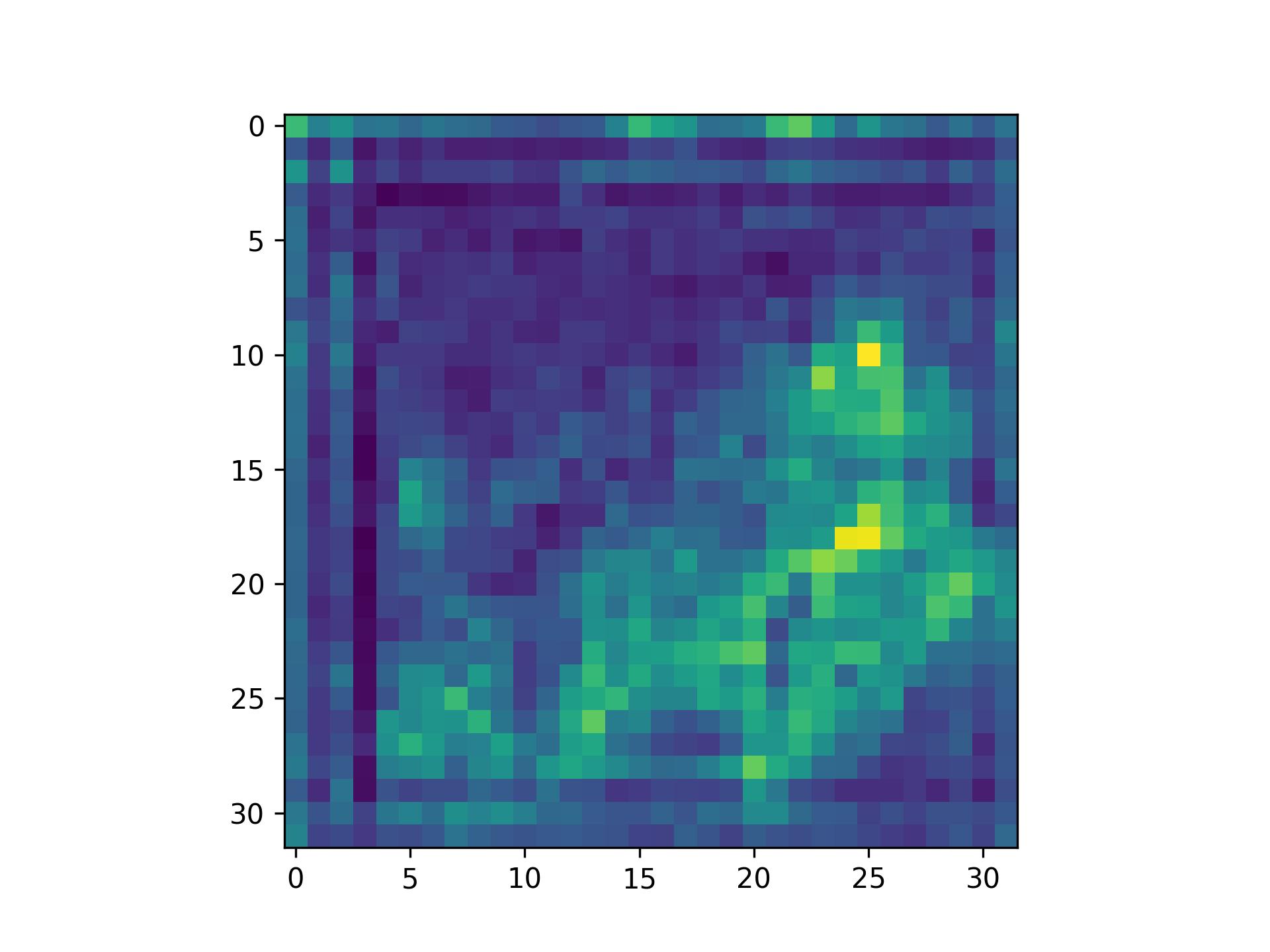}\\
  (c)
	\end{minipage}
 \begin{minipage}{0.22\linewidth}
		\centering
		\includegraphics[width=\linewidth]{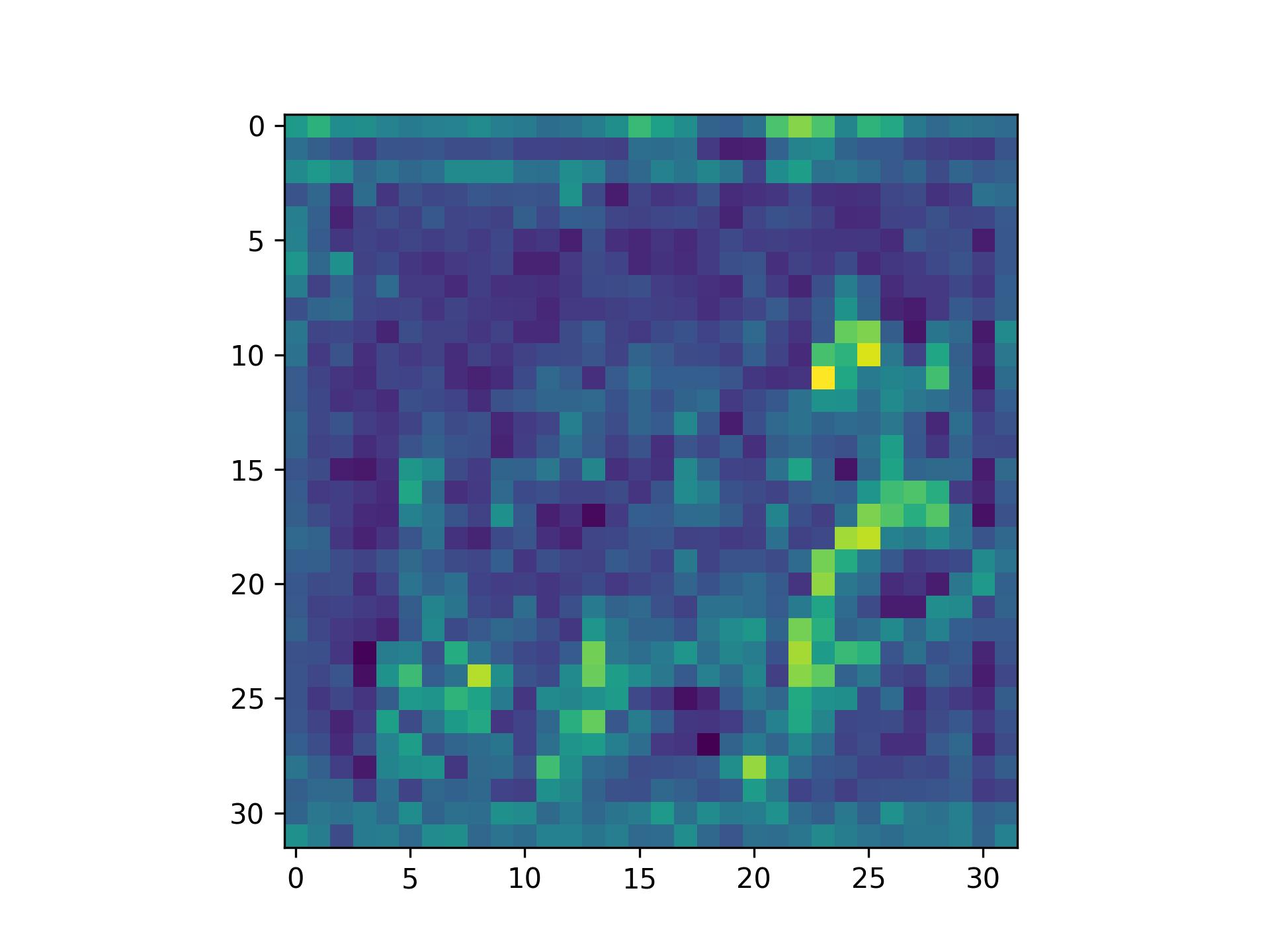}\\
  (d)
	\end{minipage}
 \caption{Feature visualization comparison, (a) and (c) are InfoPro with Patch Feature Fusion, while (b) and (d) are the original InfoPro. The feature map are obtained using ResNet-32 (K=16) as the backbone on the CIFAR-10 dataset.}
 \label{Figure 6}
\end{figure}

\noindent {\bfseries Performance Analysis of Patch Feature Fusion:} To verify the reasons for the performance improvement brought by PFF as indicated in Table \ref{Table 4}, we conduct a visualization of the features, as shown in Fig. \ref{Figure 6}. It can be observed that after adding PFF, there are more activated areas on the feature maps, indicating that the network has learned more detailed features. This demonstrates that Patch Feature Fusion (PFF), by dividing the input features into smaller patches, enables the network to concentrate on patterns that are recurrent across numerous patches, thereby further enhancing the network's performance.

Table \ref{Table 5} illustrates that PFF significantly reduces the GPU memory usage of networks incorporating the HiLo method, resulting in even lower GPU memory consumption than the original method. To understand this phenomenon, we analyze the computational complexity. Assuming we possess an input feature \(x\) with dimension \(D\) \((D = C \times H \times W)\). The auxiliary network \(f(\cdot)\) is an \(L\)-layer network with \(P\) parameters. In conventional supervised local learning methods, the entirety of \(x\) is processed through network \(f(\cdot)\) in a single pass. This results in memory consumption that includes the storage of \(x\) \((O(D))\), the storage of the auxiliary network \((O(P))\), and the storage of parameters for approximations of \(x\) in each layer \((O(L \times D))\). Consequently, the total memory usage amounts to \(O(D + P + L \times D)\). When employing PFF, the storage requirement for a single feature is reduced to \(O\left(\frac{D}{n^2}\right)\). Since the auxiliary network \(f(\cdot)\) parameters are shared for each patch, their storage remains \(O(P)\). The storage of features in the intermediate layers is reduced to \(O\left(L \times \frac{D}{n^2}\right)\). Thus, the GPU memory requirement for processing a single patch is reduced to \(O\left(\frac{D}{n^2} + P + L \times \frac{D}{n^2}\right)\). As only one patch is processed at any given moment, the overall GPU memory requirement is primarily dictated by the demands of the currently processed patch and does not increase with the number of patches.

\section{Conclusion}

This paper introduces HPFF: a novel approach that segments the network into independent and cascade levels and divides the features in the auxiliary networks into patches for computation. HiLo enables information exchange between gradient-isolated local modules, mitigating the shortsightedness issue in supervised local learning. Concurrently, PFF enhances the network's ability to capture universal features by segmenting the features in the auxiliary networks into patches, thereby conserving GPU memory. These two methods complement each other. HPFF can be seamlessly integrated into existing methods and bridging the performance gap between supervised local learning methods and BP. We validate the effectiveness of our approach across various network architectures and image classification datasets, demonstrating that our method achieves performance comparable to BP while significantly reducing GPU memory usage.

\noindent {\bfseries Limitations and Future Work:} Despite the superior performance of our proposed HPFF, the method still relies on backpropagation for loss calculation.
As suggested in {\cite{dellaferrera2022error}} and {\cite{ren2022scaling}}, methods that eliminate the need for backpropagation entirely show promise and represent one aspect of potential future exploration. 
\section*{Acknowledgement}
This work was supported by the National Natural Science Foundation of China (52278154), the Natural Science Foundation of Jiangsu (BK20231429), the Fundamental Research Funds for the Central Universities (2242024RCB0008), and assupport from the program of Zhishan Young Scholar of Southeast University.

\bibliographystyle{splncs04}
\bibliography{main}

\begin{thebibliography}{10}
\providecommand{\url}[1]{\texttt{#1}}
\providecommand{\urlprefix}{URL }
\providecommand{\doi}[1]{https://doi.org/#1}

\bibitem{bartunov2018assessing}
Bartunov, S., Santoro, A., Richards, B., Marris, L., Hinton, G.E., Lillicrap, T.: Assessing the scalability of biologically-motivated deep learning algorithms and architectures. Advances in neural information processing systems  \textbf{31} (2018)

\bibitem{belilovsky2019greedy}
Belilovsky, E., Eickenberg, M., Oyallon, E.: Greedy layerwise learning can scale to imagenet. In: International conference on machine learning. pp. 583--593. PMLR (2019)

\bibitem{bengio2006greedy}
Bengio, Y., Lamblin, P., Popovici, D., Larochelle, H.: Greedy layer-wise training of deep networks. Advances in neural information processing systems  \textbf{19} (2006)

\bibitem{coates2011analysis}
Coates, A., Ng, A., Lee, H.: An analysis of single-layer networks in unsupervised feature learning. In: Proceedings of the fourteenth international conference on artificial intelligence and statistics. pp. 215--223. JMLR Workshop and Conference Proceedings (2011)

\bibitem{crick1989recent}
Crick, F.: The recent excitement about neural networks. Nature  \textbf{337}(6203),  129--132 (1989)

\bibitem{dellaferrera2022error}
Dellaferrera, G., Kreiman, G.: Error-driven input modulation: solving the credit assignment problem without a backward pass. In: International Conference on Machine Learning. pp. 4937--4955. PMLR (2022)

\bibitem{deng2009imagenet}
Deng, J., Dong, W., Socher, R., Li, L.J., Li, K., Fei-Fei, L.: Imagenet: A large-scale hierarchical image database. In: 2009 IEEE conference on computer vision and pattern recognition. pp. 248--255. Ieee (2009)

\bibitem{dozat2016incorporating}
Dozat, T.: Incorporating nesterov momentum into adam  (2016)

\bibitem{he2016deep}
He, K., Zhang, X., Ren, S., Sun, J.: Deep residual learning for image recognition. In: Proceedings of the IEEE conference on computer vision and pattern recognition. pp. 770--778 (2016)

\bibitem{huang2017densely}
Huang, G., Liu, Z., Van Der~Maaten, L., Weinberger, K.Q.: Densely connected convolutional networks. In: Proceedings of the IEEE conference on computer vision and pattern recognition. pp. 4700--4708 (2017)

\bibitem{huo2018training}
Huo, Z., Gu, B., Huang, H.: Training neural networks using features replay. Advances in Neural Information Processing Systems  \textbf{31} (2018)

\bibitem{huo2018decoupled}
Huo, Z., Gu, B., Huang, H., et~al.: Decoupled parallel backpropagation with convergence guarantee. In: International Conference on Machine Learning. pp. 2098--2106. PMLR (2018)

\bibitem{illing2021local}
Illing, B., Ventura, J., Bellec, G., Gerstner, W.: Local plasticity rules can learn deep representations using self-supervised contrastive predictions. Advances in Neural Information Processing Systems  \textbf{34},  30365--30379 (2021)

\bibitem{jaderberg2017decoupled}
Jaderberg, M., Czarnecki, W.M., Osindero, S., Vinyals, O., Graves, A., Silver, D., Kavukcuoglu, K.: Decoupled neural interfaces using synthetic gradients. In: International conference on machine learning. pp. 1627--1635. PMLR (2017)

\bibitem{journe2022hebbian}
Journ{\'e}, A., Rodriguez, H.G., Guo, Q., Moraitis, T.: Hebbian deep learning without feedback. arXiv preprint arXiv:2209.11883  (2022)

\bibitem{keskar2017improving}
Keskar, N.S., Socher, R.: Improving generalization performance by switching from adam to sgd. arXiv preprint arXiv:1712.07628  (2017)

\bibitem{kornblith2019similarity}
Kornblith, S., Norouzi, M., Lee, H., Hinton, G.: Similarity of neural network representations revisited. In: International conference on machine learning. pp. 3519--3529. PMLR (2019)

\bibitem{krizhevsky2009learning}
Krizhevsky, A., Hinton, G., et~al.: Learning multiple layers of features from tiny images  (2009)

\bibitem{krizhevsky2012imagenet}
Krizhevsky, A., Sutskever, I., Hinton, G.E.: Imagenet classification with deep convolutional neural networks. Advances in neural information processing systems  \textbf{25} (2012)

\bibitem{le1986learning}
Le~Cun, Y.: Learning process in an asymmetric threshold network. In: Disordered systems and biological organization, pp. 233--240. Springer (1986)

\bibitem{lecun2015deep}
LeCun, Y., Bengio, Y., Hinton, G.: Deep learning. nature  \textbf{521}(7553),  436--444 (2015)

\bibitem{lee2015difference}
Lee, D.H., Zhang, S., Fischer, A., Bengio, Y.: Difference target propagation. In: Machine Learning and Knowledge Discovery in Databases: European Conference, ECML PKDD 2015, Porto, Portugal, September 7-11, 2015, Proceedings, Part I 15. pp. 498--515. Springer (2015)

\bibitem{lillicrap2014random}
Lillicrap, T.P., Cownden, D., Tweed, D.B., Akerman, C.J.: Random feedback weights support learning in deep neural networks. arXiv preprint arXiv:1411.0247  (2014)

\bibitem{lillicrap2020backpropagation}
Lillicrap, T.P., Santoro, A., Marris, L., Akerman, C.J., Hinton, G.: Backpropagation and the brain. Nature Reviews Neuroscience  \textbf{21}(6),  335--346 (2020)

\bibitem{loshchilov2016sgdr}
Loshchilov, I., Hutter, F.: Sgdr: Stochastic gradient descent with warm restarts. arXiv preprint arXiv:1608.03983  (2016)

\bibitem{lowe2019putting}
L{\"o}we, S., O'Connor, P., Veeling, B.: Putting an end to end-to-end: Gradient-isolated learning of representations. Advances in neural information processing systems  \textbf{32} (2019)

\bibitem{van2008visualizing}
Van~der Maaten, L., Hinton, G.: Visualizing data using t-sne. Journal of machine learning research  \textbf{9}(11) (2008)

\bibitem{mostafa2018deep}
Mostafa, H., Ramesh, V., Cauwenberghs, G.: Deep supervised learning using local errors. Frontiers in neuroscience  \textbf{12}, ~608 (2018)

\bibitem{netzer2011reading}
Netzer, Y., Wang, T., Coates, A., Bissacco, A., Wu, B., Ng, A.Y.: Reading digits in natural images with unsupervised feature learning  (2011)

\bibitem{nokland2016direct}
N{\o}kland, A.: Direct feedback alignment provides learning in deep neural networks. Advances in neural information processing systems  \textbf{29} (2016)

\bibitem{nokland2019training}
N{\o}kland, A., Eidnes, L.H.: Training neural networks with local error signals. In: International conference on machine learning. pp. 4839--4850. PMLR (2019)

\bibitem{ren2022scaling}
Ren, M., Kornblith, S., Liao, R., Hinton, G.: Scaling forward gradient with local losses. arXiv preprint arXiv:2210.03310  (2022)

\bibitem{rumelhart1986learning}
Rumelhart, D.E., Hinton, G.E., Williams, R.J.: Learning representations by back-propagating errors. nature  \textbf{323}(6088),  533--536 (1986)

\bibitem{rumelhart1985learning}
Rumelhart, D.E., Hinton, G.E., Williams, R.J., et~al.: Learning internal representations by error propagation (1985)

\bibitem{shen2022backpropagation}
Shen, G., Zhao, D., Zeng, Y.: Backpropagation with biologically plausible spatiotemporal adjustment for training deep spiking neural networks. Patterns  \textbf{3}(6) (2022)

\bibitem{siddiqui2023blockwise}
Siddiqui, S.A., Krueger, D., LeCun, Y., Deny, S.: Blockwise self-supervised learning at scale. arXiv preprint arXiv:2302.01647  (2023)

\bibitem{simonyan2014very}
Simonyan, K., Zisserman, A.: Very deep convolutional networks for large-scale image recognition. arXiv preprint arXiv:1409.1556  (2014)

\bibitem{wang2021revisiting}
Wang, Y., Ni, Z., Song, S., Yang, L., Huang, G.: Revisiting locally supervised learning: an alternative to end-to-end training. arXiv preprint arXiv:2101.10832  (2021)

\bibitem{xie2017aggregated}
Xie, S., Girshick, R., Doll{\'a}r, P., Tu, Z., He, K.: Aggregated residual transformations for deep neural networks. In: Proceedings of the IEEE conference on computer vision and pattern recognition. pp. 1492--1500 (2017)

\bibitem{xiong2020loco}
Xiong, Y., Ren, M., Urtasun, R.: Loco: Local contrastive representation learning. Advances in neural information processing systems  \textbf{33},  11142--11153 (2020)

\end{thebibliography}


\end{document}